\documentclass[11pt,a4paper]{article}
\usepackage[hyperref]{emnlp-ijcnlp-2019}
\usepackage{times}
\usepackage{latexsym}
\usepackage{graphicx}
\usepackage{amsmath}
\usepackage{makecell}
\usepackage{url}
\usepackage{color}
\usepackage{booktabs}
\usepackage{colortbl}
\usepackage{amssymb}
\usepackage{xspace}
\usepackage{soul}
\usepackage{CJKutf8}
\usepackage{subfigure}
\usepackage{xcolor}
\usepackage{algorithm}
\usepackage{algorithmic}
\usepackage{multirow}
\usepackage{ulem}

\newcommand{\ie}{\emph{i.e.,}\xspace}

\title{Stick to Facts: Towards Fidelity-oriented Product Description Generation}
\aclfinalcopy
\author{
Zhangming Chan$^{1,2,}$\thanks{\;\;Equal contribution. Ordering is decided by a coin flip.}$^{\;\;,}$\thanks{\;\;Contribution during internship at Alibaba Group.}$^{\;}$, 
Xiuying Chen$^{1,2,*}$, 
Yongliang Wang$^{3}$, 
Juntao Li$^{1,2}$, \\
\bf Zhiqiang Zhang$^{3}$, 
Kun Gai$^{3}$,
Dongyan Zhao$^{1,2}$ \and 
Rui Yan$^{1,2,}$\thanks{\;\;Corresponding author.} \\
$^1$ Center for Data Science, AAIS, Peking University\\
$^2$ Wangxuan Institute of Computer Technology, Peking University \\
$^3$ Alibaba Group, Beijing\\ 
{\tt \{zhangming.chan,xy-chen,ljt,zhaody,ruiyan\}@pku.edu.cn} \\
{\tt \{yongliang.wyl,zhang.zhiqiang,jingshi.gk\}@alibaba-inc.com}
}

\date{}

\begin{document}
\maketitle
\begin{abstract}
Different from other text generation tasks, in product description generation, it is of vital importance to generate faithful descriptions that stick to the product attribute information.
However, little attention has been paid to this problem.
To bridge this gap we propose a model named Fidelity-oriented Product Description Generator (FPDG).
FPDG takes the entity label of each word into account, since the product attribute information is always conveyed by entity words. 
Specifically, we first propose a Recurrent Neural Network (RNN) decoder based on the Entity-label-guided Long Short-Term Memory (ELSTM) cell, taking both the embedding and the entity label of each word as input.
Second, we establish a keyword memory that stores the entity labels as keys and keywords as values, and FPDG will attend to keywords through attending to their entity labels.
Experiments conducted a large-scale real-world product description dataset show that our model achieves the state-of-the-art performance in terms of both traditional generation metrics as well as human evaluations.
Specifically, FPDG increases the fidelity of the generated descriptions by 25\%.
\end{abstract}

\section{Introduction}
\label{sec:intro}
The effectiveness of automatic text generation has been proved in various natural language processing applications, such as neural machine translation~\cite{luong2014addressing,zhou2017neural}, dialogue generation~\cite{tao2018get,hu2019gsn}, and abstractive text summarization~\cite{timeline,Gao2019Abstractive}.
One such task, product description generation has also attracted considerable attention \cite{lipton2015capturing,wang2017statistical,zhang2019automatic}.
An accurate and attractive description not only helps customers make an informed decision but also improves the likelihood of purchase.
Concretely, the product description generation task takes several keywords describing the attributes of a product as input, and then outputs fluent and attractive sentences that highlight the feature of this product.


\begin{table}[]
\centering
\small
\resizebox{\columnwidth}{!}{
\begin{tabular}{l|l}
\hline
Input & 
\multicolumn{1}{p{6.5cm}}
        {high-waist; 
        straight; 
        jeans;
        blue;
        ZARA
        } \\ \hline
\rotatebox{270}{Bad output} & \multicolumn{1}{p{6.5cm}}
        {This \sout{UNIQLO low-waist} jeans are suitable for the curvature of the waistline and will not slip when worn for a long time. 
        The straight version and \sout{black} color together makes you even slimmer.
        } \\ \hline
\rotatebox{270}{Good output} & \multicolumn{1}{p{6.5cm}}
    {This \uline{blue jeans} look personalized and chic, making you more distinctive. 
        With a \uline{high-waist} design, the legs look even more slender, and the whole person looks taller. 
        The \uline{straight} design hides the fat in the legs and is visually slimmer.
    } \\ \hline
\end{tabular}}
\caption{Example of product description generation. The text with underline demonstrates faithful description, and text with deleteline demonstrates wrong description.}
\label{tab:intro-case}
\end{table}

There is one intrinsic difference between product description generation and other generation tasks, such as story or poem generation \cite{li2018generating,yao2019plan}; the generated description has to be faithful to the product attributes.
An example case is shown in Table~\ref{tab:intro-case}, where the good product description matches the input information, while the bad description mistakes the brand and the style of the jeans.
Our preliminary study reveals that 48\% of the outputs from a state-of-the-art sequence-to-sequence system suffer from this problem.
In the real-world e-commerce product description generation system, generating text with unfaithful information is unacceptable. 
A wrong brand name will damage the interests of advertisers, while a wrong product category might break the law by misleading consumers.
Any kind of unfaithful description will bring huge economic losses to online platforms.

Though of great importance, no attention has been paid to this problem.
Existing product description generators include \cite{chen2019towards,Zhang2019AGP}, where they focus on generating personalized descriptions and pattern-controlled descriptions, respectively.

In this paper, we address the fidelity problem in generation tasks by developing a model named \textit{Fidelity-oriented Product Description Generator} (FPDG), which takes keywords about the product attributes as inputs.
Since product attribute information is always conveyed by entity words (up to 89.72\% in input keywords), FPDG generates faithful product descriptions by taking into account the entity label of each word.
Specifically, first, we propose an Entity-label-guided Long Short-Term Memory (ELSTM) as the cell in the decoder RNN, which takes the entity label of an input word, as well as the word itself as input.
Then, we establish a keyword memory storing the input keywords with their entity labels.
In each decoding step, the current hidden state of the ELSTM focuses on proper words in this keyword memory in regard to their entity categories.
We also collect a large-scale real-world product description dataset from one of the largest e-commerce platforms in China.
Extensive experiments conducted on this dataset show that FPDG outperforms the state-of-the-art baselines in terms of traditional generation metrics and human evaluations.
Specifically, FPDG greatly improves the fidelity of generated description by 24.61\%.

 To our best knowledge, we are the first to explore the fidelity problem of product description generation.
Besides, to tackle this problem, we propose an ELSTM and a keyword memory to incorporate entity label information, so as to generate more accurate descriptions.

\section{Related Work}
\label{sec:related}

We detail related work on text generation, entity-related generation, and product description.

\textbf{Text generation.}
Recently, sequence-to-sequence (Seq2Seq) neural network models have been widely used in NLG approaches.
Their effectiveness has been demonstrated in a variety of text generation tasks, such as neural machine translation \cite{luong2014addressing,bahdanau2014neural, wu2016google}, abstractive
text summarization \cite{see2017get,hsu2018unified, chen2018fast}, dialogue generation \cite{tao2018get, xing2017topic}, etc.
Along another line, there are also works based on an attention mechanism.
\citet{vaswani2017attention} proposed a Transformer architecture that utilizes the self-attention mechanism and has achieved state-of-the-art results in neural machine translation.
Since then, the attention mechanism has been used in a variety of tasks \cite{devlin2018bert, fan2018hierarchical, zhou2018multi}.

\textbf{Entity-related generation.}
Named entity recognition (NER) is a fundamental component in language understanding and reasoning \cite{greenberg2018marginal,katiyar2018nested}.
In \cite{ji2017dynamic}, they proved that adding entity related information can improve the performance of language modeling.
Building upon this work, in \cite{clark2018neural}, they combined entity context with previous-sentence context, and demonstrated the importance of the latter in coherence test.
Another line of related work generates recipes using neural networks to track and update entity representations \cite{bosselut2017simulating}.
Different from the above works, we utilize entity labels as supplementary information to assist decoding in the text generation task.

\begin{figure*}
    \centering
    \includegraphics[scale=0.45]{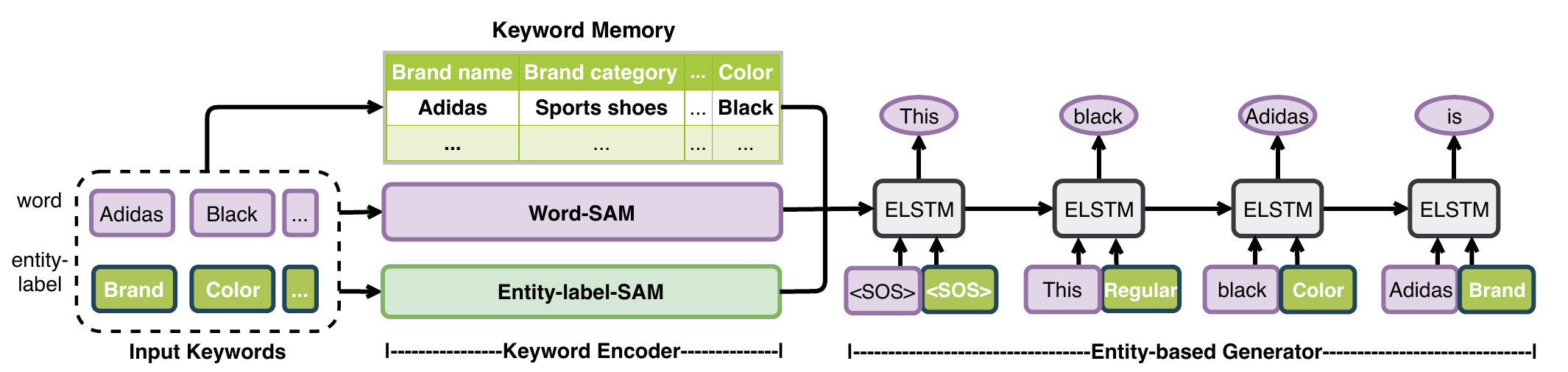}
    \caption{
        Overview of FPDG. Green denotes an entity label and purple denotes a word.
        We divide our model into two components: (1) The \textit{Keyword Encoder} stores the word and its entity label in the token memory, and uses Self-Attention Modules (SAMs) to encode words and entity labels; (2) The \textit{Entity-based Generator} generates product description based on the token memory and SAM encoders.
    }
    \label{fig:overview}
\end{figure*}

\textbf{Product descriptions.}
Quality product descriptions are critical for providing a competitive customer experience in an e-commerce platform.
Due to its importance, automatically generating the product description has attracted considerable interests.
Initial works include \cite{wang2017statistical}, which incorporates statistical methods with the template to generate product descriptions.
With the development of neural networks,  \cite{chen2019towards} explored a new way to generate personalized product descriptions by combining the power of neural networks and a knowledge base.
\cite{Zhang2019AGP} proposed a pointer-generator neural network to generate product description whose patterns are controlled.

In real-world product description generation application, however, the most important prerequisite is the fidelity of generated text, and to the best of our knowledge, no research has been conducted on this so far.

\section{Problem Formulation}
\label{sec:Problem}

FPDG takes a list of keywords $X=\{x_{1},...,x_{T_X}\}$ as inputs, where $T_X$ is the number of keywords.
These keywords are all about the important attributes of the product.
The goal of FPDG is to generate a product description $\hat{Y}=\{\hat{y}_{1},...,\hat{y}_{T_{\hat{Y}}}\}$ that is not only grammatically correct but also consistent with the input information, such as the brand name and the fashion style.
Essentially, FPDG tries to optimize the parameters to maximize the probability $P(Y|X) = \prod_{t=1}^{T_Y} P(y_t|X)$, where $Y=\{y_1,...,y_{T_Y}\}$ is the ground truth answer.

\section{Model}
\label{sec:model}

In this section, we introduce our \textit{Fidelity-oriented Product Description Generator} (FPDG) model in detail. 
An overview of FPDG is shown in Figure~\ref{fig:overview} and can be split into two modules:

(1) \textit{Keyword Encoder} (See \S~\ref{subsec:encoder}): 
We first use a key-value memory to store the entity-label as key and the corresponding word as value.
To better learn the interaction between words, we employ two self-attention modules from Transformer \cite{vaswani2017attention} to model the input keywords and their entity-labels separately.

(2) \textit{Entity-based Generator} (See \S~\ref{subsec:generator}): We propose a recurrent decoder based on Entity-label-guided LSTM (ELSTM) to generate the product description.

\subsection{Keyword Encoder}
\label{subsec:encoder}

In the product description dataset, most of the input keywords are entity words (up to 89.72\%), and the description text should be faithful to these entity words.
Hence, we incorporate entity label information to improve the accuracy of the generated text.
In this section, we introduce how to embed input keywords with entity label information.
 
To begin with, we use an embedding matrix $e$ to map a one-hot representation of each word in $x_i$ into a high-dimensional vector space.
Since our input keywords have no order information, we use the Self-Attention Module (SAM) from Transformer~\cite{vaswani2017attention} to model the temporal interactions between the words, instead of RNN-based encoder.
We use three fully-connected layers to project $e(x_i)$ into three spaces, \ie the query $q_i = F_q(e(x_i))$, the key $k_i = F_k(e(x_i))$ and the value $v_i = F_v(e(x_i))$.
The attention module then takes $q_i$ to attend to each $k_\cdot$, and uses these attention distribution results $\alpha_{i,\cdot} \in \mathbb{R}^{T_X}$ as weights to obtain the weighted sum of $v_i$, as shown in Equation~\ref{equ:transformer-sum}.
Next, we add the original word representation $e(x_i)$ on $\beta_i$ as the residential connection layer, as shown in Equation~\ref{equ:drop-add}:
\begin{align}
    \alpha_{i,j} &= \frac{\exp\left( q_i  k_j \right)}{\sum_{n=1}^{T_X} \exp\left(q_i  k_n\right)} , \label{eq:alpha}\\
    \beta_i &= \textstyle \sum_{j=1}^{T_X} \alpha_{i,j}  v_j , \label{equ:transformer-sum}\\
    \hat{h}_i &= e(x_i) + \beta_i , \label{equ:drop-add}
\end{align}
where $\alpha_{i,j}$ denotes the attention weight of the $i$-th word on the $j$-th word.
Finally, we apply a feed-forward layer on $\hat{h}_i$ to obtain the final word representation $h_i$:
\begin{align}
    h_i &= \max(0, \hat{h}_i \cdot W_1 + b_1) \cdot W_2 + b_2 , \label{eq:ffn}
\end{align}
where $W_1, W_2, b_1, b_2$ are all trainable parameters.
We refer to the above process as:
\begin{align}
     h_i = \text{SAM} \left( F_q(x_i), F_k(x_\cdot), F_v(x_\cdot) \right). \label{equ:table-encoding}
\end{align}

In the meantime, we leverage the e-commerce-adapted AliNER\footnote{\url{https://ai.aliyun.com/nlp/ner}} to label each word in the input such as ``brand name'' and ``color''.
Non-entity words are labeled as ``normal word''.
We denote $c_i$ as the entity label for the $i$-th word.
To better learn the interaction between these entity labels, we use a second SAM, taking $c_\cdot$ as input, and obtain the final representation for $c_i$ as $m_i$:
\begin{align}
     m_i = \text{SAM} \left( F_q(c_i), F_k(c_\cdot), F_v(c_\cdot) \right). 
\end{align}

We also propose a key-value keyword memory that stores the label results as shown in Figure~\ref{fig:overview}.
The keys in the keyword memory are $T_C$ representations of different entity-labels, where $T_C$ is the number of entity categories.
The values in the memory are self-attention representations of words that belong to each category.
We denote the $k$-th entity-label in the keyword memory as $c^k$, and the $i$-th word belonging to this category as $e(x^k_i)$.
This keyword memory will be incorporated into the generation process in \S\ref{subsubsec:Memory}.

\subsection{Entity-based Generator}
\label{subsec:generator}
To incorporate entity label information into generation process, we propose a modified version of LSTM, named Entity-label-guided Long Short-Term Memory (ELSTM).
We first introduce ELSTM and then introduce the RNN decoder based on ELSTM.
\begin{figure}
    \centering
    \includegraphics[scale=0.5]{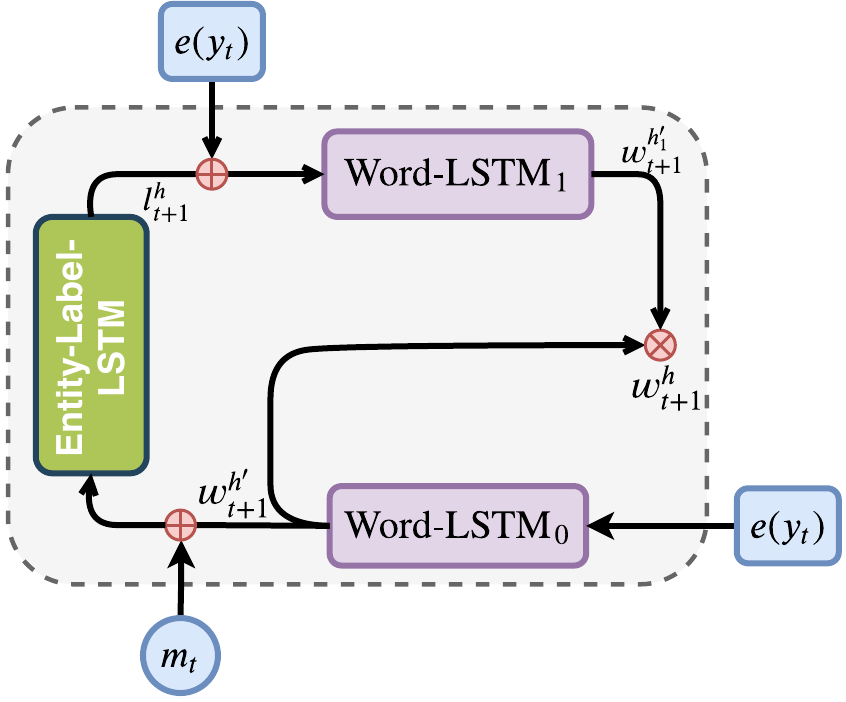}
    \caption{
        The structure of ELSTM, which is a hybrid of three LSTMs.
    }
    \vspace{-5mm}
    \label{fig:cell}
\end{figure}
\subsubsection{ELSTM}

\begin{figure}
    \centering
    \includegraphics[scale=0.5]{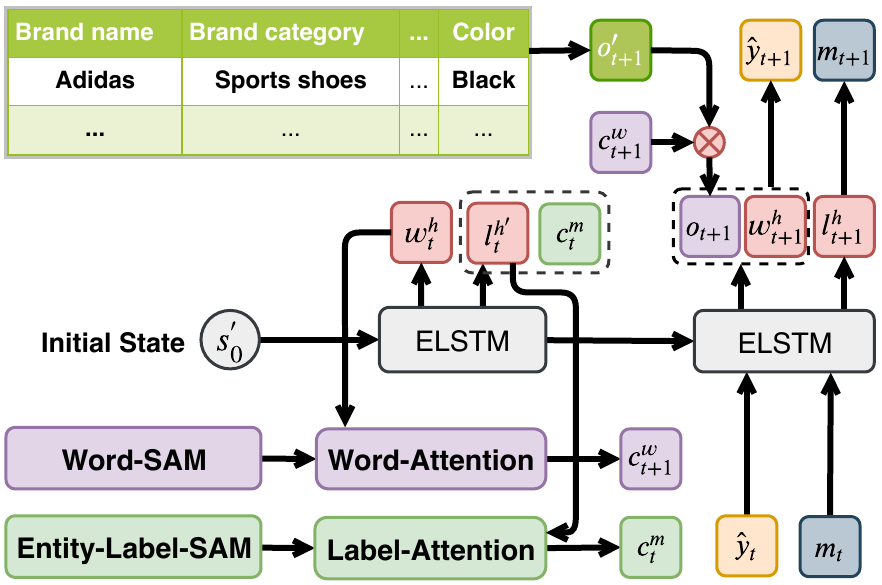}
    \caption{
        An overview of the description generator.
    }
    \label{fig:decoder}
\end{figure}

As shown in Figure~\ref{fig:decoder}, ELSTM consists of three LSTMs, two Word-LSTMs and one Entity-label-LSTM.
The hidden states of the two Word-LSTMs are integrated together, forming $w^{h}_t$, while the hidden state of Entity-Label-LSTM is $l^{h}_t$.
The word hidden state $w^{h}_t$ attends to the SAM outputs to predict the next word, while the entity label hidden state $l^{h}_t$ is used to attend to the keyword memory and predict entity label of next word.
In other words, ELSTM is a hybrid of three LSTMs with a deeper interaction between these cells.
Overall, ELSTM takes two variables as input, the embedding of an input word $e(y_t)$ and its entity label $m_t$.

The structure of each LSTM in ELSTM is the same as original LSTM, thus, is omitted here due to space limitations.
Next, we introduce the interaction in ELSTM in detail.
As shown in Figure~\ref{fig:cell}, $\text{Word-LSTM}_0$ takes the $e(y_t)$ as input, and outputs the initial word hidden state $w^{h'}_{t+1}$:
\begin{align}
    w^{h'}_{t+1}=\text{Word-LSTM}_0(w^{h}_{t},e(y_t)).
\end{align}
Next we calculate the entity label hidden state by Entity-Label-LSTM, taking both $m_{t}$ and $w^{h'}_{t+1}$ as input:
\begin{align}
    l^{h'}_{t+1} &=\text{Entity-Label-LSTM}(l^{h}_t,m_t+w^{h'}_{t+1}).
\end{align}
To further improve the accuracy of entity label prediction, we also apply another multi-layer projection, incorporating entity label context vector $c^m_t$ (introduced in \S\ref{eq:context}) to obtain polished entity hidden state $l^h_{t+1}$:
\begin{align}
    l^{h}_{t+1} &= FC([l^{h'}_{t+1}, c^m_{t}]),
\end{align}
where $[,]$ denotes the concatenation operation.

Finally, $\text{Word-LSTM}_1$ takes the entity-label hidden state $l^h_{t+1}$ and word embedding $e(y_{t})$ as input, and outputs predicted word hidden state that contains entity label information:
\begin{align}
     w^{h'_1}_{t+1}=\text{Word-LSTM}_1(w^{h}_{t},e(y_t) + l^{h}_{t+1}).
\end{align}
Note that it is possible that the predicted entity-label hidden state is not accurate enough. 
Hence, we apply a gate fusion combining $w^{h'_1}_{t+1}$ with initial word hidden state $w^{h'}_{t+1}$ to ensure the word hidden state quality:
\begin{align}
    \gamma=\sigma(FC([w^{h'_1}_{t+1},w^{h'}_{t+1}])),\\
    w^{h}_{t+1}=\gamma w^{h'_1}_{t+1}+(1-\gamma)w^{h'}_{t+1}.
    \label{gate}
\end{align}
The fusion result is $w^{h}_{t+1}$, \textit{i.e.}, the updated word hidden state.

In this way, the entity label information and word information are fully interacted in ELSTM cell, while the hidden states are updated.

\subsubsection{ELSTM-based Attention Mechanism}

Established on ELSTMs, we now have a new RNN decoder to generate descriptions, incorporating the two SAM encoders and the keyword memory.
First, we apply an Bi-RNN to encode input keywords, and use its last hidden state as decoder initial hidden states, \ie $w^h_0$ and $l^h_0$.
The $t$-th decoding step is calculated as:
\begin{align}
 \begin{bmatrix}
w^h_{t+1}\\ 
l^h_{t+1}
\end{bmatrix} = \text{ELSTM} \left(\begin{bmatrix}
w^h_t\\ 
l^h_t
\end{bmatrix}, (e(y_{t}), m_t)\right)
\label{eq:dec-step}.
\end{align}

Next, similar to the traditional attention mechanism from ~\cite{Bahdanau2015Neural}, we summarize the input word representation $e(y_{.})$ and entity label representation $m_\cdot$ into the \textit{word context vector} $c^w_{t}$ and \textit{entity label vector} $c^m_{t}$, respectively.
As for how to obtain $c^w_{t}$ and $c^m_{t}$, we use the two hidden states in ELSTM, \ie $w_t^h$ and $l_t^h$ to attend to $h_{.}$ and $m_{.}$, respectively.
Specifically, the entity label context vector $c_{t}^m$ is calculated as:
\begin{align}
\hat{\gamma}_{t,i} &= W_a \tanh \left( W_b l_t^h + W_h h_i \right), \\ 
\gamma_{t, i} &= \exp \left( \hat{\gamma}_{t, i} \right) / \textstyle \sum^{T_X}_{j=1} \exp \left(\hat{\gamma}_{t, j} \right), \label{equ:attention-sm}\\
c^m_{t} &= \textstyle \sum_{i=1}^{T_X} \gamma_{t, i} m_i. \label{eq:context}
\end{align}
The decoder state $l_t^h$ is used to attend to each entity label representation $m_i$, resulting in the attention distribution $\gamma_{t} \in \mathbb{R}^{T_X}$, as shown in Equation~\ref{equ:attention-sm}.
Then we use the attention distribution $\gamma_{t}$ to obtain a weighted sum of the entity label representations $m_\cdot$ as the word context vector $c^m_{t}$, shown in Equation~\ref{eq:context}.
$c^w_{t}$ is obtained in a similarity by using hidden state $w^h_t$ attending to $e(y_.)$, and we omit the details for brevity.
This two context vectors play different parts, and is introduced in \S\ref{project}.

\subsubsection{Incorporating Keyword Memory}
\label{subsubsec:Memory}
So far, we have finished calculating the context vector.
Next, we describe how to incorporate the guidance from the keyword memory.
We first use the entity label hidden state to attend to the entity label keys in the keyword memory by gumbel softmax~\cite{jang2016categorical} (Equation~\ref{equ:attr-key-score}), and then use the attention weights to obtain the weighted sum of self-attention representation of values in the memory (Equation~\ref{equ:entitysum}):
\begin{align}
&\pi^{k'}_t = \frac{\exp((l_t^h W_e c^k + p^k)/\tau)}{\textstyle \sum^{T_c}_{j=1} \exp((w_t^h  W_e c^j + p^k)/\tau)},\label{equ:attr-key-score} \\
&v_i^{k'} = \text{SAM} \left( F_q(e(x_i^k), F_k(e(x_{\cdot}^k), F_v(e(x_{\cdot}^k) \right), \\
&o^{'}_{t+1} = \textstyle \sum^{T_C}_{k=1} \left(\pi^k_t  \textstyle \sum^{T_{c^k}}_{i=1}(v_i^{k'})\right),\label{equ:entitysum}
\end{align}
where $p^k=-\log(-\log(g^k)), g^k \sim U(0,1),\tau \in [0,1]$ is the softmax temperature.
$T_{c_k}$ denotes the number of words in the input keywords that belong to the $k$-th entity category.
We choose gumbel-softmax instead of regular softmax because a generated word can only belong to one entity category.
In this way, FPDG first predicts the entity label of the predicted word and then uses $o^{'}_{t+1}$ to store the information of words that belong to this category.

\subsubsection{Projection Layers}
\label{project}
Next, using a fusion gate $g_t$, $o^{'}_{t+1}$ is combined with word hidden state $c^w_{t+1}$ in a similar way in Equation~\ref{gate} to obtain $o_{t+1}$.
Finally, we obtain the final generation distribution $P_{v}$ over vocabulary:
\begin{align}
&P^{v}_{t+1} = \text{softmax} \left( W_v [ o_{t+1},w^{h}_{t+1}]  + b_v \right).
\label{equ:out-proj}
\end{align}
We concatenate the memory vector $o_{t+1}$, the word context vector $c^w_{t+1}$, and the output of the decoder ELSTM $w_{t+1}^h$ as the input of the output projection layer.

Apart from predicting the next word, we also use the entity label hidden state $l^h_t$ to predict the entity label of the next word as an auxiliary task. 
The distribution $P^e_{t+1}$ over entity categories is calculated as:
\begin{align}
    P^e_{t+1}=\text{softmax}(W_e[l^{h'}_{t+1},c^m_{t+1}]+b_e)
\end{align}
We use negative log-likelihood as loss function:

\begin{align}
    \mathcal{L} &= - (\textstyle \sum^{T_{\hat{Y}}}_{t=1} \log P_{v}(y_t) + \lambda \cdot \textstyle \sum^{T_{\hat{Y}}}_{t=1} \log P_{e}(m_t)),
\label{equ:loss-fuc}
\end{align}
where $\lambda$ is the weight of entity label prediction loss.
The gradient descent method is employed to update all parameters and minimize this loss function.

\section{Experimental Setup}
\label{sec:setup}
\newcommand{\dubbelop}{$^{\blacktriangle}$}
\newcommand{\dubbelneer}{$^{\blacktriangledown}$}
\subsection{Research Questions}
We list four research questions that guide the experiments: 
\noindent \textbf{RQ1} (See \S~\ref{subsec:Overall}): What is the overall performance of FPDG?
Are generated descriptions faithful?
\noindent \textbf{RQ2} (See \S~\ref{subsec:ablation}):  What is the effect of each module in FPDG?
\noindent \textbf{RQ3} (See \S~\ref{subsec:memory}): How does the keyword memory work in each decoding step?
\noindent \textbf{RQ4} (See \S~\ref{sec:elstm}): Can ELSTM successfully capture entity label information?

\subsection{Dataset}
We collect a large-scale real-world product description dataset from one of the largest e-commerce platforms in China\footnote{\url{https://www.taobao.com/}}.
The inputs are keywords about the product attributes selected by sellers on the platform, and the product descriptions are written by professional experts that are good at marketing.
Overall, there are 404,000 training samples, and 5,000 validation and test samples.
On average, there are 10 keywords in the input, and 63 words in the product description.
89.72\% inputs are entity words.

\subsection{Evaluation Metrics}
Following \cite{fu2017aligning}, we use the evaluation package of \cite{chen2015microsoft}, which includes BLEU-1, BLEU-2, BLEU-3, BLEU-4, METEOR and ROUGE-L.
BLEU is a popular machine translation metric that analyzes the co-occurrences of n-grams between the candidate and reference sentences.
METEOR is calculated by generating an alignment between the words in the candidate and reference sentences, with an aim of 1:1 correspondence.
ROUGE-L is a metric designed to evaluate text summarization algorithms.

\cite{tao2018ruber} notes that only using the BLEU metric to evaluate text quality can be misleading. 
Therefore, we also evaluate our model by human evaluation.
Three highly educated participants are asked to score 100 randomly sampled summaries generated by Pointer-Gen and FPDG. 
We chose Pointer-Gen since its performance is relatively higher than other baselines.
The statistical significance of observed differences between the performance of two runs are tested using a two-tailed paired t-test and is denoted using \dubbelop (or \dubbelneer) for strong significance for $\alpha = 0.01$.

\subsection{Comparison Methods}

We first conduct an ablation study to prove the effectiveness of each module in FPDG.
Model w/o ELSTM is implemented with only keys in the memory because there are no entity representations as values.
Then, to evaluate the performance of our proposed dataset and model, we compare it with the following baselines:

\noindent (1) \textbf{Seq2Seq}: The sequence-to-sequence framework~\cite{Sutskever2014SequenceTS} is one of the initial works proposed for the language generation task. 

\noindent (2) \textbf{Pointer-Gen}: A sequence-to-sequence framework with pointer and coverage mechanism proposed in \cite{getto17}.

\noindent (3) \textbf{Conv-Seq2seq}: A model combining the convolutional neural networks and the sequence-to-sequence network \cite{gehring2017convolutional}.

\noindent (4) \textbf{FTSum}: A faithful summarization model proposed in \cite{cao2018faithful}, which leverages
open information extraction and dependency parse technologies to extract actual fact descriptions from the source text.

\noindent (5) \textbf{Transformer}: A network architecture that solely based on attention mechanisms, dispensing with recurrence and convolutions entirely~\cite{vaswani2017attention}.

\noindent (6) \textbf{PCPG}: A pattern-controlled product description generation model proposed in \cite{Zhang2019AGP}. We adapt the model for our scenario.

To verify whether the performance improvement is obtained by adding additional entity label inputs, we directly concatenate the word embedding with the entity embedding as input for these baselines, denoted as ``with Entity Embedding'' in Table~\ref{tab:comp_baselines}.

\begin{table*}[!ht]
\centering
\resizebox{\textwidth}{!}{
\begin{tabular}{lcccccc}
\toprule
\multirow{2}{*}{\bf Models} & \multicolumn{4}{c}{\bf BLEU Metric} & \multirow{2}{*}{\bf METEOR} & \multirow{2}{*}{\bf ROUGE-L} \\ \cmidrule{2-5}
                       & \bf BLEU-1 & \bf BLEU-2 & \bf BLEU-3 & \bf BLEU-4    &               &  \\
\midrule
Seq2seq                & 30.10    & 13.61    & 7.564    & 4.824    & 14.08   & 25.04  \\
with Entity Embedding     & 30.46    & 13.77    & 7.722    & 5.012    & 14.09        & 24.96 \\
\midrule
Pointer-Gen            & 30.66    & 13.84    & 7.933    & 5.278    & 14.10       & 24.93 \\
with Entity Embedding     & 30.92    & 13.92    & 7.988    & 5.219   & 14.14       & 24.85 \\
\midrule
Conv-Seq2seq           & 28.22    & 12.15    & 6.393    & 4.012    & 13.72    & 23.92 \\
with Entity Embedding     & 28.23    & 12.22    & 6.401    & 4.012    & 13.75    & 24.03 \\
\midrule
FTSum                  & 30.33    & 13.57    & 7.563    & 4.689    & 13.86    & 24.63 \\
with Entity Embedding     & 30.47    & 13.61    & 7.611    & 4.937    & 13.97    & 24.82 \\
\midrule
Transformer                   & 29.11    & 12.77    & 7.116    & 4.620    & 13.95    & 23.36 \\
with Entity Embedding     & 29.37    & 12.64    & 6.852    & 4.315    & 13.88    & 23.73 \\
\midrule
PCPG(including Entity) & 29.46    & 13.02    & 7.259    & 4.620    & 13.72    & 24.50 \\
\midrule
\textbf{Our FPDG}       & \bf 32.26  & \bf 14.79  & \bf 8.472  & \bf 5.629  & \bf 15.17  & \bf 25.27 \\
\midrule
\midrule
w/o KW-MEM           & 31.85    & 14.36    & 8.134    & 5.351    & 15.01 & 25.09 \\
w/o ELSTM              & 31.41    & 14.57    & 8.430    & 5.630    & 14.98    &  25.25 \\
\bottomrule
\end{tabular}
}
\caption{RQ1: Comparison between baselines.}
\label{tab:comp_baselines}
\end{table*}

\subsection{Implementation Details}

We implement our experiments in Pytorch\footnote{\url{https://pytorch.org/}} on Tesla V100 GPUs\footnote{Our model has been incorporated into Chuangyi Taobao (\url{https://chuangyi.taobao.com/pages/smartPhrase}) and deployed online for the Double 11 Shopping Festival.}. 
The word and entity-label embedding dimensions are set to 256.
The number of hidden units and the entity-label hidden size are also set to 256.
All inputs were padded with zeros to a maximum keyword number of the batch.
There are 36 categories of entity labels together.
$\lambda$ is set to 0 in the first 500 steps, and 0.6 in the rest of the training process.
We performed minibatch cross-entropy training with a batch size of 256 documents for 15 training epochs.
we set the minimum encoding step to 15 and maximum decoding step size to 70.
During decoding, we employ beam search with a beam size of 4 to generate a more fluent sentence.
It took around 6 hrs on GPUs for training.
After each epoch, we evaluated our model on the validation set and chose the best performing model for the test set. 
We use the Adam optimizer~\cite{Duchi2010AdaptiveSM} as our optimizing algorithm and the learning rate is 1e-3.

\section{Experimental Results}
\subsection{Overall Performance}
\label{subsec:Overall}

For research question \textbf{RQ1}, we examine the performance of our model and baselines in terms of BLEU, as shown in Table~\ref{tab:comp_baselines}.
Firstly, among all baselines, Pointer-Gen obtains the best performance, outperforming the worst baseline Conv-Seq2Seq by 2.44 in BLEU-1.
Secondly, directly concatenating the entity label embedding with the word embedding does not bring much help, only leading to an improvement of 0.26 in BLEU-1 for the Pointer-Gen model.
Finally, our model outperforms all baselines for all metrics, outperforming the strongest baseline Pointer-Gen, by 5.22\%, 6.86\%, 6.79\%, and 6.65\% in terms of BLEU-1, BLEU-2, BLEU-3, and BLEU-4, respectively.

\begin{table}[t]
    \centering
    \newcommand{\cbkgrnd}{\cellcolor{blue!15}}
    \begin{tabular}{@{}lccc@{}}
        \toprule
        & Fluency & Informativity & Fidelity \\ 
        \midrule
        Pointer-Gen & 2.23\phantom{0} & 1.84\phantom{0}&  1.91\phantom{0}\\
        FPDG & \textbf{2.46}\dubbelop & \textbf{2.19}\dubbelop & \textbf{2.38}\dubbelop\\
        \bottomrule
    \end{tabular}
    \label{tab:comp_human_baslines}
    \caption{RQ1: Human evaluation comparison with Pointer-Gen baseline.}
\end{table}

As for human evaluation, we ask three highly educated participants to rank generated summaries in terms of fluency, informativity, and fidelity.
The rating score ranges from 1 to 3 with 3 being the best.
The results are shown in Table 3, where FPDG outperforms Pointer-Gen by 10.31\% and 19.02\% in terms of fluency and informativity, and, specifically, FPDG greatly improves the fidelity value by 24.61\%.
We also conduct the paired student t-test between our model and Pointer-Gen, and the result demonstrates the significance of the above results. 
The kappa statistics are 0.35 and 0.49, respectively, which indicates fair and moderate agreement between annotators\footnote{\cite{landis1977measurement} characterize kappa values $<$ 0 as no agreement, 0-0.20 as slight, 0.21-0.40 as fair, 0.41-0.60 as moderate, 0.61-0.80 as substantial, and 0.81-1 as almost perfect agreement.}.

\subsection{Ablation Study}
\label{subsec:ablation}

Next, we turn to research question \textbf{RQ2}.
We conduct ablation tests on the usage of the keyword memory and ELSTM, corresponding to FPDG w/o KW-MEM and ELSTM, respectively.
The ROUGE score result is shown at the bottom of  Table~\ref{tab:comp_baselines}.
Performances of all ablation models are worse than that of FPDG in terms of almost all metrics, which demonstrates the necessity of each module in FPDG.
Specifically, ELSTM makes the greatest contribution to FPDG, improving the BLEU-1, BLEU-2 scores by 2.71\% and 1.51\%.

\begin{figure}
    \centering
    \subfigure{ 
        \includegraphics[clip,width=1\columnwidth]{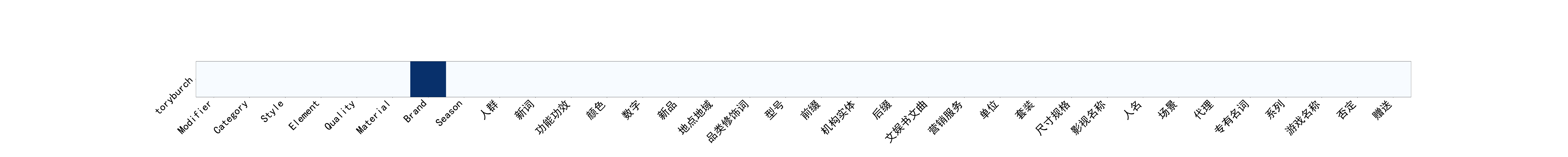}}
    \subfigure{
        \includegraphics[clip,width=1\columnwidth]{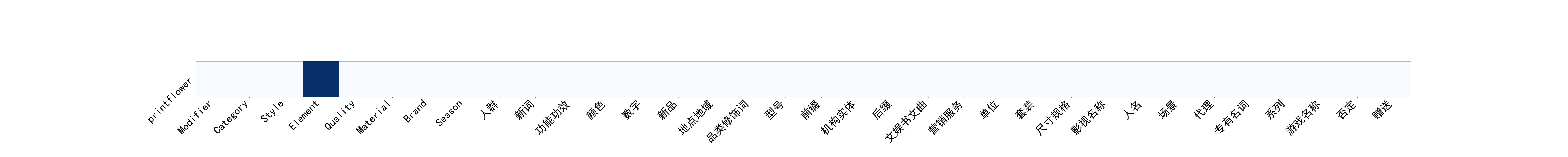}}
    \caption{RQ3: Visualizations of entity-label-attention when generating the word on the left.}
    \label{fig:attention}
\end{figure}

\subsection{Analysis of Keyword Memory}
\label{subsec:memory}

We then address \textbf{RQ3} by analyzing the entity-label-level attention on the keyword memory.
Two representative cases are shown in Figure~\ref{fig:attention}.
The figure in the above is the attention map when generating the word ``toryburch'', and the bottom figure is when generating the word ``printflower''.
The darker the color, the higher the attention.
Due to limited space, we omit irrelevant entity categories.
When generating ``toryburch'', which is a brand name, the entity-label-level attention pays most attention to the ``Brand'' entity label, and when generating ``flower'', which is a style element, it mostly pays attention to ``Element''.
This example demonstrates the effectiveness of the entity-label-level attention.

\subsection{Analysis of ELSTM}
\label{sec:elstm}

We now turn to \textbf{RQ4}; whether or not the ELSTM can capture entity label information.
We examine this question by verifying whether ELSTM can predict the entity label of the generated word.
The accuracy of the predicted entity label is calculated in a teacher-forcing style, \ie each ELSTM takes the ground truth entity label and word as input, and outputs the entity label of the next word.
We employ recall at position $k$ in $n$ candidates ($R_n@k$) as evaluation metrics.
Over the whole test dataset, $R_1@36$ is 64.12\%, $R_2@36$ is 80.86\%, and $R_3@36$ is 94.02\%, which means ELSTM can capture the entity label information to a great extent and guide the word generation.

\begin{CJK*}{UTF8}{gkai}
 \label{tab:case}
    \begin{table}[t]
    \centering
        \small
        \begin{tabular}{l|l}
            \toprule
          \multicolumn{2}{p{7.2cm}}{Input:正品; Prada;/;普拉达; 男包;  尼龙; 双肩背包; 男士; 休闲; 旅行包; 电脑包; 登山包 (authentic; Prada; man bag; leisure; travel bag; laptop bag; backpack)}  \\
            \hline
            \rotatebox{270}{Reference}                       & \multicolumn{1}{p{6.5cm}}{这款来自prada的拉链双肩包。采用纯色设计，包面上的字母印花图案十分显眼，令人眼前一亮，瞬间脱颖而出。开合双拉链处理，使用顺滑，方便拿取物品。大容量的包包，为物品的放置提供足够的空间。(This zippered shoulder bag from Prada. With a solid color design, the letter print on the surface of the bag is very conspicuous, making it instantly stand out. Open and close double zipper makes it easy to use take items smoothly. Large-capacity bag provides enough space for the placement of items.)
          }   \\ \hline
          \rotatebox{270}{Pointer-Gen}                         & \multicolumn{1}{p{6.5cm}}{这款来自prada的翻盖双肩包。采用了翻盖的设计，具有很好的防盗效果，让你的出行更加安全可靠。大容量的包型设计，满足日常的生活所需，让你的出行更加方便快捷。(This flipover bag from Prada. The \sout{clamshell design has a good anti-theft effect, making your trip safer and more reliable}. The large-capacity package design meets the needs of everyday life, making your travel more convenient and quick.}    \\ \hline
            \rotatebox{270}{FPDG}                           & \multicolumn{1}{p{6.5cm}}{这款来自prada的纯色双肩包。纯色的设计，给人一种大气又干练的感觉，打造时下流行的简约风格。拉链的开合设计，划拉顺畅，增加物品安全性。大容量的包型，既能容纳生活物品，还让携带更加的便捷。(This solid color backpack from Prada. The solid color design gives people an atmosphere and a sense of exquisiteness, creating a simple style that is popular nowadays. \uline{The opening and closing design of the zipper smoothes the stroke} and increases the safety of the item. Large-capacity package can accommodate living items and make it easier to carry.} \\ 
            \bottomrule
        \end{tabular}
        \caption{Examples of the generated answers by Pointer-Gen and FPDG.
        The text with underline demonstrates faithful description, and text with deleteline demonstrates wrong description.}
    \end{table}
\end{CJK*}

We also show a case study in Table 4.
The description generated by Pointer-Gen introduces the Prada bag as a ``clamshell bag has a good anti-theft effect'', which is contrary to the fact.
While our model generates the faithful description: ``The opening and closing design of the zipper smoothes the stroke''.

\section{Conclusion and Future Work}
In this paper, we explore the fidelity problem in product description generation.
To tackle this challenge, based on the consideration that product attribute information is typically conveyed by entity words, we incorporate the entity label information of each word to enable the model to have a better understanding of the words and better focus on key information.
Specifically, we propose an Entity-label-guided Long Short-Term Memory (ELSTM) and a token memory to store and capture the entity label information of each word.
Our model outperforms state-of-the-art methods in terms of BLEU and human evaluations by a large margin.
In the near future, we aim to fully prevent the generation of unfaithful descriptions and bring FPDG online.

\section*{Acknowledgments}
We would like to thank the reviewers for their constructive comments.
This work was supported by the National Key Research and Development Program of China (No. 2017YFC0804001), the National Science Foundation of China (NSFC No. 61876196 and NSFC No. 61672058). Rui Yan was sponsored by Alibaba Innovative Research (AIR) Grant. Rui Yan is the corresponding author.

\bibliography{emnlp-ijcnlp-2019}
\bibliographystyle{acl_natbib}
\end{document}